\documentclass{article}

\usepackage{settings/spconf}
\ninept
\usepackage[english]{babel}

\usepackage{type1cm} 
\usepackage{graphicx} 
\usepackage{xspace} 
\usepackage{balance} 
\usepackage{booktabs} 
\usepackage{multirow} 
\usepackage[font={bf}, tableposition=top]{caption} 
\usepackage{subcaption} 
\usepackage{bold-extra} 
\usepackage{algorithm}
\usepackage{algorithmic}
\usepackage{microtype} 
\usepackage{siunitx} 
\usepackage{xfrac} 
\usepackage{amsmath}
\usepackage{amssymb}
\usepackage{mathtools}
\usepackage{amsfonts}
\usepackage{amsthm}
\usepackage{bbm}
\usepackage[hyphens]{url} 
\usepackage[bookmarks, pdftex, colorlinks=true]{hyperref} 
\usepackage{cleveref} 
\usepackage{csquotes}
\usepackage{bm}
\usepackage{xspace}
\usepackage{physics}
\usepackage{newunicodechar}
\newunicodechar{⟓}{\ensuremath{\uplus}}
\usepackage{tikz}
\usetikzlibrary{positioning}
\usetikzlibrary{bayesnet}
\usetikzlibrary{arrows}
\usetikzlibrary{shapes}
\usetikzlibrary{fit}
\usepackage{tikz-cd}
\usepackage{bbold}
\usetikzlibrary{calc, positioning, arrows.meta, shapes.geometric, fit, backgrounds}
\usepackage{pgf}
\usepackage{tikz-3dplot}
\usepackage{tcolorbox}
\usepackage{pgfplots}
\pgfplotsset{compat=1.18}

\newcommand{\spara}[1]{\smallskip\noindent\textbf{#1}}

\usepackage[dvipsnames]{xcolor}
\hypersetup{
   bookmarks, pdftex,
   colorlinks=true,
   pagebackref=true, backref=page,
   linkcolor={red!50!black},
   filecolor={green!50!black},
   citecolor={green!50!black}, 
   urlcolor={blue!80!black},
}

\graphicspath{{fig}}

\theoremstyle{plain}
\newtheorem{theorem}{Theorem}

\newtheorem{definition}{Definition}

\newcommand{\reall}{\ensuremath{\mathbb{R}}\xspace}

\newcommand{\ort}[1]{\ensuremath{\mathrm{O}({#1})}\xspace}

\newcommand{\frob}[1]{\ensuremath{\norm{#1}_{\mathrm{F}}}\xspace}

\newcommand{\F}{\ensuremath{\mathbf{F}}\xspace}

\newcommand{\w}{\ensuremath{\mathbf{w}}\xspace}

\newcommand{\Omatrix}{\ensuremath{\mathbf{O}}\xspace}

\definecolor{mypurple}{RGB}{254, 68, 218}

\newcommand{\Kronecker}{\ensuremath{\mathbb{K}}\xspace}

\newcommand{\bdO}{\ensuremath{\mathbb{O}}\xspace}

\title{Learning the structure of Connection Graphs}

\name{Leonardo Di Nino$^{1,2}$, Gabriele D'Acunto$^{1,2}$, Sergio Barbarossa$^{1}$, and Paolo Di Lorenzo$^{1,2}$
\thanks{The work was supported by the SNS JU project 6G-GOALS \cite{strinati2024goal} under the EU’s Horizon program Grant Agreement No 101139232, by Huawei Technology France SASU under Grant N. Tg20250616041, and by the  European Union under the Italian National Recovery and Resilience Plan of NextGenerationEU, partnership on Telecommunications of the Future (PE00000001- program RESTART).}}

\address{$^1$ DIET Department, Sapienza University of Rome, Via Eudossiana 18, 00184, Rome, Italy\\ 
$^2$ National Inter-University Consortium for Telecommunications (CNIT), Parma, Italy\\
{E-mail: \{leonardo.dinino, gabriele.dacunto,  sergio.barbarossa, paolo.dilorenzo\}@uniroma1.it}
}
\begin{document}

\maketitle

\begin{abstract}
Connection graphs (CGs) extend traditional graph models by coupling network topology with orthogonal transformations, enabling the representation of global geometric consistency. They play a key role in applications such as synchronization, Riemannian signal processing, and neural sheaf diffusion. In this work, we address the inverse problem of learning CGs directly from observed signals. We propose a principled framework based on maximum pseudo-likelihood under a consistency assumption, which enforces spectral properties linking the connection Laplacian to the underlying combinatorial Laplacian. Based on this formulation, we introduce the Structured Connection Graph Learning (SCGL) algorithm, a block-optimization procedure over Riemannian manifolds that jointly infers network topology, edge weights, and geometric structure. Our experiments show that SCGL consistently outperforms existing baselines in both topological recovery and geometric fidelity, while remaining computationally efficient.

\end{abstract}


\section{Introduction}\label{sec:introduction}

Graph signal processing (GSP) \cite{sandryhaila2013discrete} has established as a powerful framework to analyze data supported on irregular domains. 
Key to GSP is the graph shift operator---typically the graph Laplacian \cite{smola2003kernels}---whose algebraic and spectral properties enable diffusion processes on graphs. 
This machinery serves as the foundation for both classical convolutional methods \cite{leus2023graph} and graph-based deep learning architectures \cite{wu2020comprehensive}.
Consequently, the learning of graphs and related shift operators from data is a fundamental problem for signal processing over unknown networks.
Due to identification issues, graph learning is typically guided by prior knowledge of properties matching the observed signal.  
Numerous methods, ranging from signal processing and probabilistic modeling approaches \cite{mateos2019connecting} to deep learning techniques \cite{Kazi_2023} have contributed to graph learning.
They mainly pursue two approaches:
\emph{(i)} enforcing a desired topology, balancing sparsity with connectivity,
and \emph{(ii)} promoting signal smoothness, i.e., low variation of the signal across connected nodes on the learned graph.
These approaches determine graph Laplacians with desirable spectral properties; however, both are affected by the inherent limitation of GSP:
they capture only local, pairwise interactions among nodes, limiting the ability to model global consistency of the signal across the network. 
In this pursuit, sheaf-theoretic approaches \cite{curry2014sheaves,ayzenberg2025sheaf} enrich graph-based representations by assigning structured data to nodes and edges, and relating them via structure preserving maps, thus defining a very general functorial framework that can be specialized according to the underlying process or representation task.
Thanks to their ability to glue local information into global states, network sheaves allow to capture a wider range of network processes moving beyond graph diffusion \cite{bodnar2023neuralsheafdiffusiontopological}.
In these models structural consistency inherently shapes the sheaf Laplacians \cite{hansen2020laplacians} which no longer depend only on the underlying graph topology but also on the functorial nature of network sheaves.
This makes the data-driven learning of sheaf Laplacians a fundamental, open challenge calling for new coupling criteria between inferred systems and observed signals beyond graph smoothness and connectivity. 
Within this family of models, connection graphs (CGs) associate nodes and edges with inner product vector spaces, while encoding their relationships through structure-preserving maps represented by orthogonal linear transformations. 
Over the past, CGs have emerged for their relevance in synchronization problems \cite{bandeira2013}, Riemannian signal processing \cite{battiloro2024tangent}, and neural sheaf diffusion \cite{pmlr-v196-barbero22a}. Motivated by these applications, we investigate the problem of learning CGs from data.

\noindent\textbf{Related works.} An established framework is Vector Diffusion Maps (VDM) \cite{singer2012vector}, which approximate the CG Laplacian through geometric principles and perform nonlinear dimensionality reduction for manifold learning. 
Conversely, we approach the problem from an inverse perspective, extending structured graph learning \cite{kumar2020unified} to CGs by incorporating the notion of consistency \cite{chung2014ranking}—a sufficient condition for desirable spectral properties of CGs and a principle that also underlies (flat) bundle neural networks \cite{bamberger2024bundle}. 
Our methodology aligns with the smooth graph learning paradigm \cite{dong2016learning}, already extended to sheaf Laplacians in \cite{8683709} via semidefinite programming. However, the method proposed in the latter work can properly recover some classes of sheaves but not the CG, which is inherently linked to the non-Euclidean geometry of the orthogonal manifold. Finally, in \cite{10942997}, we addressed the limits of conic programming for sheaf learning using Procrustes alignment and binary edge sampling.

\noindent\textbf{Contributions.}  In this work, we introduce a principled framework for learning connection Laplacians from observed data under the assumption of consistency. 
Our main contributions are as follows:
\emph{(i)} we formulate a maximum pseudo-likelihood problem that extends spectral control from graphs to CGs, steering the learning process toward meaningful network structures while ensuring geometric consistency; 
\emph{(ii)} we develop an iterative algorithm, termed Structured Connection Graph Learning (SCGL), which leverages block descent optimization on Riemannian manifolds to jointly recover both topological and geometric patterns; 
\emph{(iii)} we demonstrate the effectiveness of SCGL through synthetic experiments on random and geometric graphs, showing significant improvements over existing baselines in CG learning. 
Together, these contributions establish SCGL as a versatile tool for geometry-aware network topology inference.
\section{Preliminary definitions}

In this section, we briefly introduce CGs. 
For a detailed discussion of the mapping between CGs and cellular sheaves, refer to \cite{curry2014sheaves, hansen2020laplacians}.

\begin{definition}[Connection graph]
    Given a graph $\mathcal{G}\coloneq(V,E)$, a CG $\mathbb{G} = \langle \mathcal{G}, \mathbb{R}^n,\w, \ort{n} \rangle$ is specified by a copy of a real-valued n-dimensional Euclidean vector space $\mathbb{R}^n$ at each node $v \in V$, a non negative weight $\mathbf{w}_e$ for each edge $e \in E$ and an orthogonal matrix $\mathbf{O}_e \in \ort{n}:=\{\mathbf{A}\in\mathbb{R}^{n\times n}:\mathbf{A}^\top =\mathbf{A}^{-1}\}$ for each edge $e \in E$.
\end{definition}

\begin{definition}[Connection Laplacian] Let $\mathbb{G} = \langle \mathcal{G}, \mathbb{R}^n,\w, \ort{n} \rangle$ be a CG with $v$ nodes. The connection Laplacian $\mathbb{L} \in \mathbb{R}^{vn \times vn}$ is defined as a block matrix satisfying $[\mathbb{L}]_{ij}=\mathbf{0}$ for each $(i,j) \notin E$, and satisfying for each edge $(i,j) \in E$:
\begin{equation}\label{eq:KLapl}
        [\mathbb{L}]_{ij}  \coloneqq \begin{cases}
            -w_{ij} \mathbf{O}_{ij} \quad & i>j\,,\\
            [\mathbb{L}]_{ji}^\top & i<j\,,\\
            \sum_{j\neq i} w_{ij}\mathbb{I}_n\quad & i=j\,.
        \end{cases}
    \end{equation}
\end{definition}
\noindent CGs enrich a graph by defining how local vector-valued information over nodes coherently glue to global states via orthogonal transformations accordingly to an underlying geometry reflected by the connection Laplacian. 
The spectral properties of this Laplacian operator, in turn, reveal rich insights into the structure of the CG and the nature of diffusion processes defined on it \cite{sharp2019vector}.
In particular, the kernel of the connection Laplacian encodes \textit{synchronization} \cite{bandeira2013}, modeling richer configurations beyond constant states associated to the kernel of the combinatorial Laplacian, thus moving beyond \textit{consensus}.
Differently from graphs, topology alone cannot determine the spectral properties of a CG \cite{chung1997spectral}: 
the spectral theory of connection Laplacians necessarily involves the linear maps over the edges and their inherent geometry.
We say that a CG is \textit{consistent} when the orthogonal maps over edges compose to the identity along each cycle of the graph.
This in turn yields a useful spectral characterization: 
\begin{theorem}[\cite{chung2014ranking}]\label{th:consistency_G}
    Let $\mathbb{G}=\langle \mathcal{G}, \mathbb{R}^n,\mathbf{w}, \ort{n} \rangle$ be a CG having $v$ nodes and connection Laplacian $\mathbb{L} \in \reall^{vn \times vn}$.
    Let $\mathbf{L}$ be the combinatorial Laplacian of the underlying graph $\mathcal{G}$.
    Hence, the following are equivalent:
    \begin{enumerate}
        \item\label{th:consistency_G_1} $\mathbb{G}$ is consistent; 
        \item\label{th:consistency_G_2}The eigenvalues of $\mathbb{L}$ are the eigenvalues of $\mathbf{L}$, each of multiplicity $n$;
        \item\label{th:consistency_G_3} For each node $i$ in $G$, we can find $\Omatrix_i \in \mathrm{SO}(n) :=\{\mathbf{A}\in \ort{n}:\mathrm{det}(\mathbf{A})=1\}$ such that for any edge $(i,j) \in E$, we have $\Omatrix_{ij} = \Omatrix_i^\top \Omatrix_j$. 
    \end{enumerate}
\end{theorem}
\noindent \Cref{th:consistency_G}
 states that for a consistent CG the spectrum $\mathbf{\Gamma}$ of the connection Laplacian is tied with the spectrum $\mathbf{\Lambda}$ of the combinatorial Laplacian as $\mathbf{\Gamma} = (\mathbf{\Lambda} \otimes \mathbb{I}_n)$, where $\otimes$ denote Kronecker product. 
Therefore, given a connected graph---which always possesses an eigenvalue in zero---Theorem \ref{th:consistency_G} yields a connection Laplacian with an $n$-dimensional kernel, linking the combinatorial Laplacian $\mathbf{L}$ and the connection Laplacian $\mathbb{L}$ via orthogonal matrices over the nodes factorizing the edge maps.

\section{Problem formulation}

We assume that our dataset is given by $\mathbf{X} = \{\mathbf{x}_1,...,\mathbf{x}_M\}$, where each signal realization is represented by 
\(\mathbf{x}_i \in \mathbb{R}^{nv}\), obtained by stacking the local node measurements 
\(\mathbf{x}_v \in \mathbb{R}^n, \; v \in V\). 

We model our signals as following a Gaussian distribution $\mathcal{N}_{vn}(0,\mathbb{L}^\dagger)$, where $\mathbb{L}^\dagger$ is the pseudoinverse of the connection Laplacian to be inferred.
Under this model, the estimation of the connection Laplacian from observed data can be done via maximum pseudo-likelihood posing the following regularized minimum total variation problem:
\begin{equation}\label{eq:P1}
    \underset{\mathbb{L} \in \mathcal{CL}}{\mathrm{min}} \ - \log\mathrm{gdet}(\mathbb{L})+ \Tr(\mathbf{S}\mathbb{L})  \, .
    \tag{P1}
\end{equation}
Here $\mathbf{S} = \frac{1}{M-1}\mathbf{X}\mathbf{X}^\top$ is the empirical covariance matrix, and $\mathcal{CL}$ is a proper feasible set for connection Laplacians. Under consistency, Theorem \ref{th:consistency_G} yields the following key identities:
\begin{equation}\label{eq:lap_comb}
    \bdO\mathbb{L}\bdO^\top = \mathbf{L} \otimes \mathbb{I}_n \, \Leftrightarrow \mathbb{L} = \bdO^\top(\mathbf{L} \otimes \mathbb{I}_n)\bdO\,,
\end{equation}
where \( \bdO = \mathrm{blkdiag}(\{\mathbf{O}_v\}_{v \in V}) \) is the block-diagonal matrix collecting the local orthogonal basis \( \mathbf{O}_v \) at each node \( v \).
Leveraging \eqref{eq:lap_comb}, \eqref{eq:P1} reduces to an optimization in the combinatorial Laplacian $\mathbf{L}$ and the block-diagonal basis matrix $\bdO$, over the feasible sets of graph Laplacians $\mathcal{GL}$ and $\mathcal{O}$, respectively:
\begin{equation}\label{eq:P2}
    \underset{
  \substack{
    \mathbf{L} \in \mathcal{GL} \\
    \bdO \in \mathcal{O}
  }
}{\mathrm{min}} \ - \log \mathrm{gdet}\{\bdO^\top(\mathbf{L} \otimes \mathbb{I}_n)\bdO\}+\Tr\{\mathbf{S}\bdO^\top(\mathbf{L} \otimes \mathbb{I}_n)\bdO\}  \, .
    \tag{P2}
\end{equation}
\noindent In \eqref{eq:P2}, the explicit definition of the admissible set of graph Laplacians $\mathcal{GL}$ would be required.
Instead, following the approach of \cite{kumar2020unified}, we reformulate the problem by introducing a linear operator that maps edge weights $\mathbf{w}$ directly to a Kronecker–structured Laplacian of the form $\mathbf{L}\otimes \mathbb{I}_n$. 
This will allow us to optimize over the edge weights rather than over the Laplacian matrices themselves.

\begin{definition}[Kronecker-structured  Laplacian operator]\label{def:KLapl}
    The linear operator $\mathcal{L}_\Kronecker:\reall^{v (v-1)/2} \rightarrow \reall^{vn \times vn}$ reads block-wise, for $d_j = -j + \frac{j-1}{2}(2v-j)$, as:
        \begin{equation}\label{eq:KLapl}
        [\mathcal{L}_\Kronecker(\mathbf{w})]_{ij}  \coloneqq \begin{cases}
            -\mathbf{w}_{i+d_j} \mathbb{I}_n \quad & i>j\,,\\
            [\mathcal{L}_\Kronecker(\mathbf{w})]_{ji} & i<j\,,\\
            -\sum_{j\neq i} [\mathcal{L}_\Kronecker(\mathbf{w})]_{ij}\quad & i=j\,.
        \end{cases}
    \end{equation}
\end{definition}
\noindent Using \eqref{eq:KLapl}, \Cref{eq:lap_comb} writes as:
\begin{equation}\label{eq:consistency_1}
    \mathbb{L} = \mathbb{O}^\top\mathcal{L}_\Kronecker(\w)\mathbb{O} \, .
\end{equation}
Furthermore, under Theorem \ref{th:consistency_G}, we can introduce the eigendecomposition of $\mathbb{L}$ with respect to its own set of eigenvectors $\mathbf{U}$ and to the eigenvalues $\mathbf{\Lambda}$ of the combinatorial Laplacian:  

\begin{equation}\label{eq:consistency_2}
    \mathbb{L} = \mathbf{U}\mathbf{\mathbf{\Gamma}} \mathbf{U}^\top = \mathbf{U}(\mathbf{\Lambda} \otimes \mathbb{I}_n)\mathbf{U}^\top \,.
\end{equation}
Thus, using \Cref{eq:consistency_1,eq:consistency_2}, we get
\begin{equation}\label{eq:consistency_3}
\mathbb{O}^\top\mathcal{L}_\Kronecker(\w)\mathbb{O} = \mathbf{U}(\mathbf{\Lambda} \otimes \mathbb{I}_n)\mathbf{U}^\top\,.
\end{equation}
We incorporate constraint \eqref{eq:consistency_3} into the objective of \eqref{eq:P2} through Lagrangian relaxation, while introducing a regularizer $\Psi(\mathbf{w})$ to promote sparsity in the edge weights. 
The resulting problem reads as:

\begin{align}\label{eq:P3}
\min_{\bdO,\, \mathbf{w},\, \mathbf{U},\, \mathbf{\Lambda}} \quad & 
    -n \log\mathrm{gdet}(\mathbf{\Lambda}) + \Tr\left\{ \mathbf{S} \bdO^\top \, \mathcal{L}_\Kronecker(\mathbf{w})\bdO\right\} \nonumber \\
& + \alpha\, \Psi(\mathbf{w}) \nonumber \\
&  + \frac{\beta}{2} \frob{\bdO^\top\mathcal{L}_\Kronecker(\mathbf{w})\bdO
- \mathbf{U} (\mathbf{\Lambda} \otimes \mathbb{I}_n) \mathbf{U}^\top}^2\nonumber \\
\text{subject to} \quad & \mathbf{w} \geq 0, \tag{P3} \\
& \mathbf{U}^\top \mathbf{U} = \mathbb{I}_{nv}, \nonumber \\
& \bdO = \mathrm{blkdiag}(\{\mathbf{O}_v\}_{v \in V}), \nonumber \\
& \mathbf{O}_v^\top \mathbf{O}_v = \mathbb{I}_n \quad \text{for all } v \in V, \nonumber \\
& \mathbf{\Lambda} = \mathrm{diag}(\mathbf{\lambda}) \in \mathcal{S}_{\mathbf{\Lambda}}, \nonumber 
\end{align}
where $\alpha,\beta \geq 0$ are regularization parameters, while $\mathcal{S}_{\mathbf{\Lambda}}$ denotes the set of spectral constraints imposed on $\mathbf{L}$—for instance, enforcing $\lambda_1=0$ to ensure graph connectivity. Interestingly, solving \eqref{eq:P3} allows the joint learning of \emph{(i)} the combinatorial graph via structured graph learning \cite{kumar2020unified}, with spectral constraints enforcing topological priors; and \emph{(ii)} the geometry of the connection Laplacian, including the associated orthogonal transformations, under CG consistency.

\section{Structured Connection Graph Learning}
Problem \eqref{eq:P3} is non-convex, both because of the objective function itself and due to the orthonormality constraints on the eigenvectors $\mathbf{U}$ and the node bases $\bdO$. 
Nevertheless, we design an algorithmic solution based on alternating block minimization and Riemannian optimization, which efficiently handles the manifold constraints and guarantees convergence to stationary points. 
For brevity, we omit the convergence proof, which follows \cite{kumar2020unified}.
In the sequel, we provide the main steps of our \textbf{S}tructured \textbf{C}onnection \textbf{G}raph \textbf{L}earning (SCGL) algorithm. 
The detailed derivations are omitted due to lack of space.

\spara{Block update in $\mathbf{w}$.} 
We update $\mathbf{w}$ via Minorization-Maximization (MM)\cite{7547360}. 
Linearization around the current iterate $\mathbf{w}^t$ yields:
\begin{equation}\label{eq:w_update}
\mathbf{w}^{t+1} = \mathcal{P}^+_{\frac{\alpha}{\beta}\Psi} \Big\{ \mathbf{w}^t -
\frac{1}{\tau} \mathcal{L}_\Kronecker^* \Big[ f(\mathbf{w}^t) \Big] \Big\}\,, 
\end{equation}
where
$ \displaystyle f(\mathbf{w}^t) = \mathcal{L}_\Kronecker(\mathbf{w}^t) 
- \bdO \left[
\mathbf{U}(\mathbf{\Lambda} \otimes \mathbb{I}_n) \mathbf{U}^\top 
- \frac{1}{\beta} \mathbf{S}
\right] \bdO^\top\,.$\\
Here, $\mathcal{P}^+_{\frac{\alpha}{\beta}\Psi}$ is the proximal operator induced by $\frac{\alpha}{\beta}\Psi$ projected onto the positive orthant, and $\tau = 2nv$ is the Lipschitz constant of $\mathcal{L}_\Kronecker^*\mathcal{L}_\Kronecker$.
The adjoint operator $\mathcal{L}_\Kronecker^*$ is defined with respect to the standard inner product applied to \eqref{eq:KLapl} \cite{kumar2020unified}: given $\mathbf{Y} \in \mathbb{R}^{nv \times nv}$ with block entries $\mathbf{Y}_{ij} \in \mathbb{R}^{n \times n}$, the $k$-th component of $\mathcal{L}_\Kronecker^*(\mathbf{Y}) \in \mathbb{R}^{v(v-1)/2}$, with $i > j$ and $k = i - j + \frac{(j-1)}{2}(2v - j)$, is:
\begin{equation}
[\mathcal{L}_\Kronecker^*(\mathbf{Y})]_k = \Tr\left( \mathbf{Y}_{ii} + \mathbf{Y}_{jj} - \mathbf{Y}_{ij} - \mathbf{Y}_{ji} \right)\,.
\end{equation}

\spara{Block update in $\mathbf{O}$.} The learning problem in $\mathbf{O}$ is
\begin{equation}\label{eq:O_update}
    \underset{\bdO\in \mathcal{O}}{\mathrm{min}} \ \Tr\{\bdO \mathbf{S}\bdO^\top \mathcal{L}_\Kronecker(\w)\} + \frac{\beta}{2}\frob{\mathcal{L}_\Kronecker(\w)-\bdO\mathbf{U}(\mathbf{\Lambda}\otimes\mathbb{I}_n)\mathbf{U}^\top\bdO^\top}^2\,.
\end{equation}
We solve this step numerically via Riemannian gradient descent \cite{absil2009optimization} over the product manifold $\mathrm{SO}(n)^v$.

\spara{Block update in $\mathbf{U}$.} For a connected graph, the subproblem in $\mathbf{U}$ reduces to an eigenvalue problem on the Stiefel manifold $\mathrm{St}(nv, n(v-1))$ of orthonormal matrices in $\mathbb{R}^{nv \times n(v-1)}$: 

\begin{equation}
    \underset{\mathbf{U}\in\mathrm{St}(nv,n(v-1))}{\mathrm{max}} \Tr\{\mathbf{U}^\top\bdO^\top\mathcal{L}_\Kronecker(\w)\bdO\mathbf{U}(\mathbf{\Lambda} \otimes\mathbb{I}_n)\}\,.
\end{equation}
This problem is solved by the $n(v-1)$ principal eigenvectors associated to non-zero eigenvalues of the current estimate of the connection Laplacian \cite{absil2009optimization}:
\begin{equation}\label{eq:U_update}
    \mathbf{U}^{t+1} = [\mathrm{Eigenvecs}\{\bdO^\top\mathcal{L}_\Kronecker(\w)\bdO\}]_{:,(n+1):}\,.
\end{equation}

\spara{Block update in $\mathbf{\Lambda}$.} Assuming a connected graph, the problem with respect to $\mathbf{\Lambda} = \mathrm{diag}(\lambda)$ is an isotonic regression where we can replace the generalized log-determinant with the standard log-determinant under a proper re-indexing:
\begin{equation}\label{eq:IsoReg}
    \underset{c_1\leq\mathbf{\lambda}_2 \leq ... \leq \mathbf{\lambda}_v\leq c_2}{\mathrm{min}} -n\sum_{i=1}^{v-1}\log(\mathbf{\lambda}_{i+1}) + \sum_{i=1}^{v-1}\frac{\beta}{2}\frob{\mathbf{M}_{ii}-\mathbf{\lambda}_{i+1}\mathbb{I}_n}^2\,,
\end{equation}
where $c_1,c_2$ are constants used to avoid degenerate spectral behaviours, $\mathbf{M} = \mathbf{U}^\top \bdO^\top \mathcal{L}_\Kronecker(\mathbf{w}) \bdO \mathbf{U}$, and $\mathbf{M}_{ii}$ are its diagonal blocks. 
The isotonic regression algorithm from \cite{kumar2020unified} is guaranteed to converge to the solution of \Cref{eq:IsoReg} in $V-1$ iterations, once initialized in the KKT solutions of the problem:
\begin{equation}\label{eq:LAMBDA_KKT}
\mathbf{\lambda}_{i+1} = \frac{1}{2n} \left[\Tr(\mathbf{M}_{ii}) + \sqrt{\Tr(\mathbf{M}_{ii})^2 + \frac{4n^2}{\beta}} \right], \ i = 1,...v-1\,.
\end{equation}

\spara{Computational cost.} The complexity of the algorithm is dominated by the optimization steps in $\bdO$ and in $\mathbf{U}$, which are both of complexity $O(V^3n^3)$ due to the involved eigendecompositions.
However, this means that it does not get any worse with respect to structured graph learning \cite{kumar2020unified}, which also has the update step in $\mathbf{U}$, except for the scaling effect due to the dimension of the vector space over the nodes.
Moreover, compared to the conic programming approach in \cite{8683709}, SCGL achieves a more efficient parametrization of the feasible set: under the consistency assumption, $\bdO$ factorizes edge maps into node bases, reducing the space complexity from $O(V^2n^2)$ to $O(Vn^2)$.
\begin{figure*}[!t]
  \centering
  \includegraphics[width=0.91\linewidth]{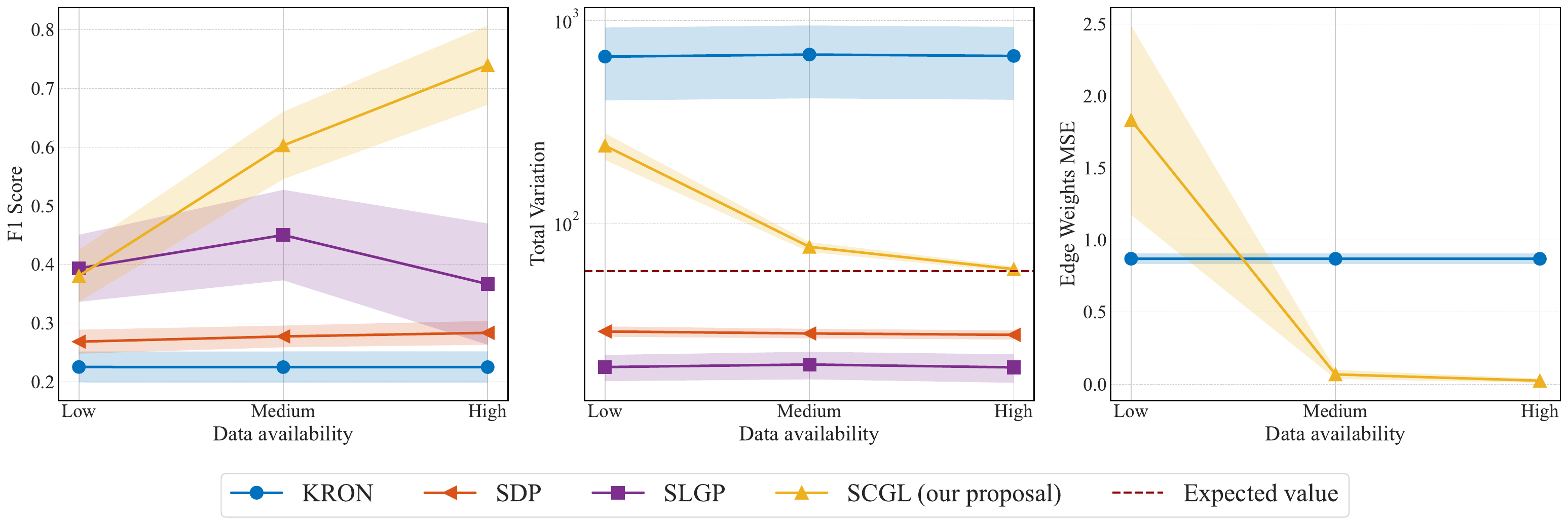}
  \caption{F1 score (left), empirical total variation (central) and MSE on edge weights (right) for the identification of ER CGs.}
  \label{fig:experiment_1a}
\end{figure*}
\begin{figure*}[!t]
  \centering
  \includegraphics[width=0.9\linewidth]{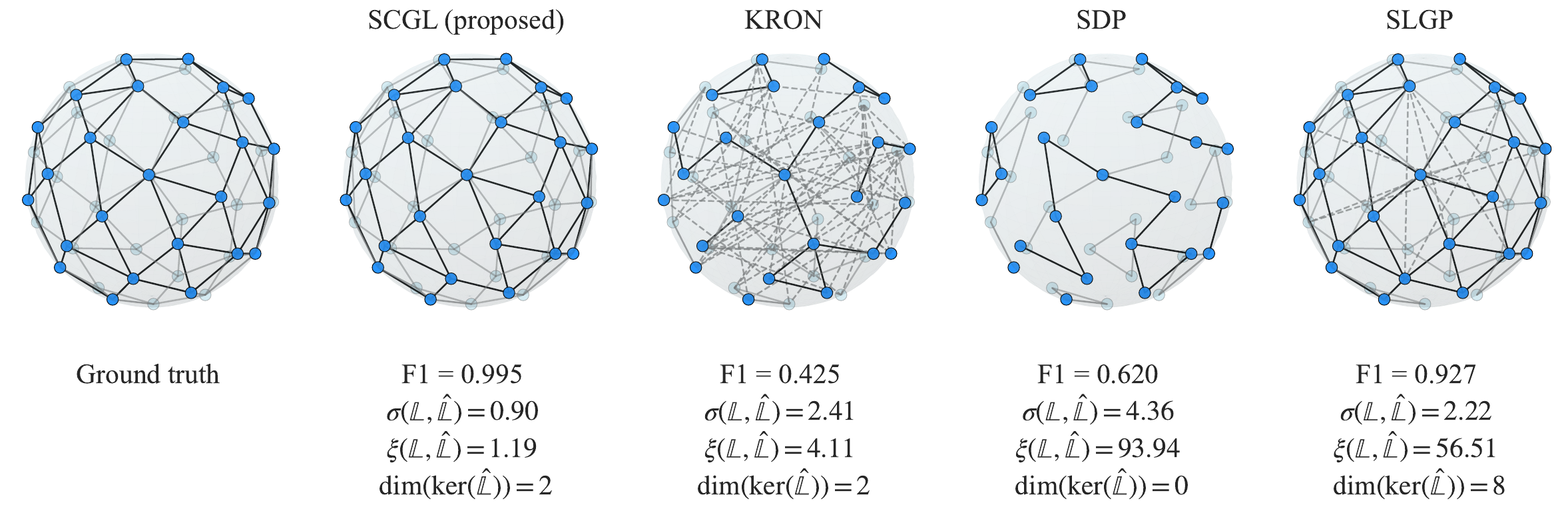}
  \caption{Graphs retrieved with SCGL and the considered baselines from data generated around a discrete sphere whose ground true connection Laplacian is derived from VDM and spectral synchronization. \vspace{-.2cm}}
  \label{fig:experiment_1b}
\end{figure*}

\section{Numerical results}
We evaluate our SCGL method through synthetic experiments in CG learning; specifically, we focus on
\emph{(i)} random Erdős–Rényi (ER) CGs,
and \emph{(ii)} geometric 
CGs over discretized spheres.

We consider three baselines.
First, \textit{Kronecker Laplacian} (KRON)--our SCGL with fixed local bases $\mathbf{O}_v = \mathbb{I}_n$ for all $v \in V$-- to quantify the benefits of optimizing the node maps over \ort{n}.
Second, \textit{Smooth learning via SDP} (SDP), the semi-definite programming formulation from \cite{8683709} restricted to Laplacians with diagonal blocks proportional to the identity.
SDP hinges on signal smoothness; however, it uses an enlarged feasible set corresponding to a convex cone including the connection Laplacians.
Hence, SDP is not guaranteed to recover a connection Laplacian as it does not match the CG Riemannian geometry.
Last, \textit{Smooth learning with geometric priors} (SLGP), our previous work \cite{10942997} which solves a minimum total variation problem with known edge set cardinality and maps constrained to \ort{n}.
Hyperparameters are validated via cross-validation and learned Laplacians undergo no post-processing.

\spara{ER CGs}.
We generate consistent CGs over an ER graph with $|V| = 30$ nodes and edge probability $p_{\mathrm{ER}} = 1.1 \frac{\log V}{V}$, enforcing connectivity by random wiring if necessary. 
Node bases are sampled from $\mathrm{SO}(2)$ and edge maps constructed to satisfy consistency, with weights drawn from $\mathrm{Unif}(0.2, 3)$. 
We draw $M$ training samples from $\mathcal{N}(0, \mathbb{L}^{\dagger})$, where $\mathbb{L}$ is the ground-truth connection Laplacian, and metrics are averaged over 20 trials. We define three data availability scenarios based on the sampling ratio $r =\frac{M}{2|V|}$, namely \textit{low} for $r = 1.5$, \textit{medium} for $r = 5$, and \textit{high} for $r = 15$. 

A thorough evaluation of the methods entails jointly measuring their topological and geometric inference performance. 
The former is measured by the F1 score on the recovered sparsity pattern: 
for SCGL and KRON we also report the MSE in recovering the edge weights, being the only two methods capable to learn them. 
Regarding the latter, direct entry-wise comparison is ill-posed since orthonormal bases are identifiable only up to a rotation matrix.
Therefore, we compute the empirical total variation $\hat{\mathcal{E}}_\mathbb{L}(\mathbf{Y}) = M^{-1} \Tr(\hat{\mathbb{L}} \mathbf{Y}\mathbf{Y}^T)$ on a test set $\mathbf{Y}$.
Beyond serving as a smoothness proxy, this measure should asymptotically coincide with the theoretical value $\mathbb{E}_{\mathbf{X}}[M^{-1}\Tr(\mathbb{L}\mathbf{X}\mathbf{X}^T)]= \mathrm{rank}(\mathbb{L})$, which reflects the true system geometry. 
Any deviation indicates inconsistency. 

\Cref{fig:experiment_1a} shows the results.
In a regime of data scarcity, all methods fail, with SLGP and SCGL performing similarly. 
Conversely, as the sample size increases, SCGL outperforms the baselines, both in identifying the sparsity pattern of the network (left) and in approximating the true connection Laplacian (center). 
Additionally, SCGL is accurate in the edge weights estimation (right).
Thus, most false positive edges are assigned negligible weights.

\spara{Spherical CGs.} 
We consider a sphere in $\mathbb{R}^3$ discretized using a Fibonacci lattice \cite{stanley1975fibonacci} with 50 points, then we build a k-NN graph over it with $k = 4$ as our reference structure.
The connection Laplacian is then approximated using Vector Diffusion Maps (VDM, \cite{singer2012vector})--we adopt the convention on the sign of the operators to be positive semi-definite--and we subsequently retrieve a consistent connection Laplacian via spectral synchronization \cite{d2021ranking}.
Using the latter in the generative model above, we sample $M=2000$ snapshots.
We then test all the methods in learning the ground-truth connection Laplacian.
For each estimated connection Laplacian $\hat{\mathbb{L}}$ we considered the following metrics: 
\emph{(i)} the F1 score for sparsity pattern; 
\emph{(ii)} the average spectral distance \cite{JOVANOVIC20121425} 
\begin{equation}
    \sigma(\mathbb{L},\hat{\mathbb{L}}) = \frac{1}{nv}\sum_{i=1}^{nv} |\mathbf{\Lambda}_i(\mathbb{L}) - \mathbf{\Lambda}_i(\hat{\mathbb{L}})|\,;    
\end{equation}
\emph{(iii)} the integrated heat diffusion distance \cite{hammond2013graph} 
\begin{equation}
    \xi(\mathbb{L}, \hat{\mathbb{L}}) = \underset{T \to \infty}{\lim}\frac{1}{T} \int_0^{T} \big\| e^{-t \mathbb{L}} - e^{-t \hat{\mathbb{L}}} \big\|_\F^2 \, dt
\,.    
\end{equation}
\Cref{fig:experiment_1b} shows the results.
Notably, SCGL not only recovers the underlying graph with minimal error (as indicated by the F1 score), but also achieves the lowest spectral and heat diffusion distances among all methods. 
At the same time, it preserves consistency, as evidenced by the correctly retrieved kernel dimension.
Overall, these results demonstrate that SCGL accurately reconstructs both the topology and geometry of the original connection graph.

\vspace{-.1cm}
\section{Conclusions}
\vspace{-.1cm}
In this paper, we introduced a framework for learning consistent CGs from observed data, extending structured graph learning to the more expressive setting of sheaf-like structures. The proposed SCGL algorithm formulates the task as a maximum pseudo-likelihood problem with spectral constraints and solves it via block coordinate descent with Riemannian optimization. 
This enables the joint recovery of the combinatorial graph, edge weights, and orthogonal transformations defining the connection Laplacian.
Experiments on Erdős–Rényi and geometric CGs show that SCGL consistently outperforms baselines in recovering network topology and geometry.
These results confirm SCGL as an efficient and accurate tool for geometry-aware graph learning, bridging topology, geometry, and signal processing. Future work includes extending SCGL to handle noise and model violations, incorporating flexible topological and geometric priors, addressing inconsistent CGs and validating the framework on real-world data.

\balance
\bibliographystyle{settings/bibstile}
\bibliography{settings/bibliography}

\begin{thebibliography}{10}

\bibitem{strinati2024goal}
Emilio~Calvanese Strinati, Paolo Di~Lorenzo, Vincenzo Sciancalepore, Adnan Aijaz, Marios Kountouris, Deniz G{\"u}nd{\"u}z, Petar Popovski, Mohamed Sana, Photios~A Stavrou, Beatriz Soret, et~al.,
\newblock ``Goal-oriented and semantic communication in 6{G} {AI}-native networks: The 6{G}-{GOALS} approach,''
\newblock in {\em 2024 Joint European Conference on Networks and Communications \& 6G Summit (EuCNC/6G Summit)}. IEEE, 2024, pp. 1--6.

\bibitem{sandryhaila2013discrete}
Aliaksei Sandryhaila and Jos{\'e}~MF Moura,
\newblock ``Discrete signal processing on graphs,''
\newblock {\em IEEE transactions on signal processing}, vol. 61, no. 7, pp. 1644--1656, 2013.

\bibitem{smola2003kernels}
Alexander~J Smola and Risi Kondor,
\newblock ``Kernels and regularization on graphs,''
\newblock in {\em Learning Theory and Kernel Machines: 16th Annual Conference on Learning Theory and 7th Kernel Workshop, COLT/Kernel 2003, Washington, DC, USA, August 24-27, 2003. Proceedings}. Springer, 2003, pp. 144--158.

\bibitem{leus2023graph}
Geert Leus, Antonio~G Marques, Jos{\'e}~MF Moura, Antonio Ortega, and David~I Shuman,
\newblock ``Graph signal processing: History, development, impact, and outlook,''
\newblock {\em IEEE Signal Processing Magazine}, vol. 40, no. 4, pp. 49--60, 2023.

\bibitem{wu2020comprehensive}
Zonghan Wu, Shirui Pan, Fengwen Chen, Guodong Long, Chengqi Zhang, and S~Yu Philip,
\newblock ``A comprehensive survey on graph neural networks,''
\newblock {\em IEEE transactions on neural networks and learning systems}, vol. 32, no. 1, pp. 4--24, 2020.

\bibitem{mateos2019connecting}
Gonzalo Mateos, Santiago Segarra, Antonio~G Marques, and Alejandro Ribeiro,
\newblock ``Connecting the dots: Identifying network structure via graph signal processing,''
\newblock {\em IEEE Signal Processing Magazine}, vol. 36, no. 3, pp. 16--43, 2019.

\bibitem{Kazi_2023}
Anees Kazi, Luca Cosmo, Seyed-Ahmad Ahmadi, Nassir Navab, and Michael~M. Bronstein,
\newblock ``Differentiable {G}raph {M}odule ({DGM}) for graph convolutional networks,''
\newblock {\em IEEE Transactions on Pattern Analysis and Machine Intelligence}, vol. 45, no. 2, pp. 1606–1617, Feb. 2023.

\bibitem{curry2014sheaves}
Justin~Michael Curry,
\newblock {\em Sheaves, cosheaves and applications},
\newblock University of Pennsylvania, 2014.

\bibitem{ayzenberg2025sheaf}
Anton Ayzenberg, Thomas Gebhart, German Magai, and Grigory Solomadin,
\newblock ``Sheaf theory: from deep geometry to deep learning,''
\newblock {\em arXiv e-prints}, pp. arXiv--2502, 2025.

\bibitem{bodnar2023neuralsheafdiffusiontopological}
Cristian Bodnar, Francesco Di~Giovanni, Benjamin Chamberlain, Pietro Lio, and Michael Bronstein,
\newblock ``Neural sheaf diffusion: A topological perspective on heterophily and oversmoothing in {GNN}s,''
\newblock {\em Advances in Neural Information Processing Systems}, vol. 35, pp. 18527--18541, 2022.

\bibitem{hansen2020laplacians}
Jakob Hansen,
\newblock {\em Laplacians of Cellular Sheaves: Theory and Applications},
\newblock Ph.D. thesis, University of Pennsylvania, 2020.

\bibitem{bandeira2013}
Afonso~S. Bandeira, Amit Singer, and Daniel~A. Spielman,
\newblock ``A cheeger inequality for the graph connection laplacian,''
\newblock {\em SIAM Journal on Matrix Analysis and Applications}, vol. 34, no. 4, pp. 1611--1630, 2013.

\bibitem{battiloro2024tangent}
Claudio Battiloro, Zhiyang Wang, Hans Riess, Paolo Di~Lorenzo, and Alejandro Ribeiro,
\newblock ``Tangent bundle convolutional learning: from manifolds to cellular sheaves and back,''
\newblock {\em IEEE Transactions on Signal Processing}, 2024.

\bibitem{pmlr-v196-barbero22a}
Federico Barbero, Cristian Bodnar, Haitz S\'aez~de Oc\'ariz~Borde, Michael Bronstein, Petar Veli\v{c}kovi\'c, and Pietro Li\`o,
\newblock ``Sheaf {N}eural {N}etworks with {C}onnection {L}aplacians,''
\newblock in {\em Proceedings of Topological, Algebraic, and Geometric Learning Workshops 2022}, 2022, pp. 28--36.

\bibitem{singer2012vector}
Amit Singer and H-T Wu,
\newblock ``Vector diffusion maps and the connection {L}aplacian,''
\newblock {\em Communications on pure and applied mathematics}, vol. 65, no. 8, pp. 1067--1144, 2012.

\bibitem{kumar2020unified}
Sandeep Kumar, Jiaxi Ying, Jos{\'e} Vin{\'\i}cius de~M Cardoso, and Daniel~P Palomar,
\newblock ``A unified framework for structured graph learning via spectral constraints,''
\newblock {\em Journal of Machine Learning Research}, vol. 21, no. 22, pp. 1--60, 2020.

\bibitem{chung2014ranking}
Fan Chung, Wenbo Zhao, and Mark Kempton,
\newblock ``Ranking and sparsifying a connection graph,''
\newblock {\em Internet Mathematics}, vol. 10, no. 1-2, pp. 87--115, 2014.

\bibitem{bamberger2024bundle}
Jacob Bamberger, Federico Barbero, Xiaowen Dong, and Michael~M Bronstein,
\newblock ``Bundle {N}eural {N}etworks for message diffusion on graphs,''
\newblock in {\em ICML 2024 Workshop on Geometry-grounded Representation Learning and Generative Modeling}.

\bibitem{dong2016learning}
Xiaowen Dong, Dorina Thanou, Pascal Frossard, and Pierre Vandergheynst,
\newblock ``Learning {L}aplacian matrix in smooth graph signal representations,''
\newblock {\em IEEE Transactions on Signal Processing}, vol. 64, no. 23, pp. 6160--6173, 2016.

\bibitem{8683709}
Jakob Hansen and Robert Ghrist,
\newblock ``Learning {S}heaf {L}aplacians from {S}mooth {S}ignals,''
\newblock in {\em ICASSP 2019 - 2019 IEEE International Conference on Acoustics, Speech and Signal Processing (ICASSP)}, 2019, pp. 5446--5450.

\bibitem{10942997}
Leonardo Di~Nino, Sergio Barbarossa, and Paolo Di~Lorenzo,
\newblock ``Learning {S}heaf {L}aplacian {O}ptimizing {R}estriction {M}aps,''
\newblock in {\em 2024 58th Asilomar Conference on Signals, Systems, and Computers}, 2024, pp. 59--63.

\bibitem{sharp2019vector}
Nicholas Sharp, Yousuf Soliman, and Keenan Crane,
\newblock ``The vector heat method,''
\newblock {\em ACM Transactions on Graphics (TOG)}, vol. 38, no. 3, pp. 1--19, 2019.

\bibitem{chung1997spectral}
Fan~RK Chung,
\newblock {\em Spectral graph theory}, vol.~92,
\newblock American Mathematical Soc., 1997.

\bibitem{7547360}
Ying Sun, Prabhu Babu, and Daniel~P. Palomar,
\newblock ``Majorization-{M}inimization {A}lgorithms in {S}ignal {P}rocessing, {C}ommunications, and {M}achine {L}earning,''
\newblock {\em IEEE Transactions on Signal Processing}, vol. 65, no. 3, pp. 794--816, 2017.

\bibitem{absil2009optimization}
P-A Absil, Robert Mahony, and Rodolphe Sepulchre,
\newblock ``Optimization algorithms on matrix manifolds,''
\newblock Princeton University Press, 2009.

\bibitem{stanley1975fibonacci}
Richard~P Stanley,
\newblock ``The {F}ibonacci lattice,''
\newblock {\em The Fibonacci Quarterly}, vol. 13, no. 3, pp. 215--232, 1975.

\bibitem{d2021ranking}
Alexandre d'Aspremont, Mihai Cucuringu, and Hemant Tyagi,
\newblock ``Ranking and synchronization from pairwise measurements via {SVD},''
\newblock {\em Journal of Machine Learning Research}, vol. 22, no. 19, pp. 1--63, 2021.

\bibitem{JOVANOVIC20121425}
Irena Jovanović and Zoran Stanić,
\newblock ``Spectral distances of graphs,''
\newblock {\em Linear Algebra and its Applications}, vol. 436, no. 5, pp. 1425--1435, 2012.

\bibitem{hammond2013graph}
David~K Hammond, Yaniv Gur, and Chris~R Johnson,
\newblock ``Graph diffusion distance: A difference measure for weighted graphs based on the graph laplacian exponential kernel,''
\newblock in {\em 2013 IEEE global conference on signal and information processing}. IEEE, 2013, pp. 419--422.

\end{thebibliography}

\end{document}